\pdfoutput=1

\documentclass[11pt]{article}

\usepackage{EACL2023}

\usepackage{times}
\usepackage{latexsym}
\usepackage[T1]{fontenc}
\usepackage[utf8]{inputenc}
\usepackage{tabularx,booktabs}

\usepackage{microtype}

\usepackage{inconsolata}
\usepackage{graphicx}
\usepackage{eurosym}
\usepackage{tablefootnote}
\graphicspath{ {images/} }
\usepackage{hyperref}
\usepackage{textcomp}
\usepackage{hhline}
\usepackage[T1]{fontenc}
\usepackage{listings}
\usepackage{stfloats}
\usepackage{csquotes}
\usepackage{arydshln}
\usepackage{booktabs}
\usepackage{array}
\usepackage[super]{nth}
\usepackage[inline,shortlabels]{enumitem}
\usepackage{multirow}
\usepackage{makecell}
\usepackage{siunitx}
\usepackage{multicol}
\usepackage{svg}
\usepackage{float}
\usepackage{amssymb}
\usepackage{amsmath,amsfonts,amssymb}
\usepackage[caption=false]{subfig}
\usepackage{graphicx}

\title{AdapterSoup: Weight Averaging to Improve Generalization of Pretrained Language Models}

\author{ 
	Alexandra Chronopoulou$^{\star\dag\triangledown}$ \quad Matthew E. Peters$^{\ddagger}$ \quad Alexander Fraser$^{\triangledown}$ \quad Jesse Dodge$^{\star \ddagger}$\\
	$^{\triangledown} $Center for Information and Language Processing, LMU Munich, Germany \\
     $^{\triangledown}$Munich Center for Machine Learning, Germany \\
	$^{\ddagger}$Allen Institute for Artificial Intelligence, Seattle, WA \\
	}
\begin{document}
\maketitle
\begin{abstract}

Pretrained language models (PLMs) are trained on massive corpora, but often need to specialize to specific domains. A parameter-efficient adaptation method suggests training an adapter for each domain on the task of language modeling. This leads to good in-domain scores but can be impractical for domain- or resource-restricted settings. A solution is to use a related-domain adapter for the novel domain at test time. In this paper, we introduce \textit{AdapterSoup}, an approach that performs weight-space averaging of adapters trained on \emph{different} domains. Our approach 
is embarrassingly parallel: first, we train a set of domain-specific adapters; then, for each novel domain, we determine which adapters should be averaged at test time. 
We present extensive experiments showing that AdapterSoup consistently improves performance to new domains without extra training. We also explore weight averaging of adapters trained on the \emph{same} domain with different hyper-parameters, and show that it preserves the performance of a PLM on new domains while obtaining strong in-domain results. We explore various approaches for choosing which adapters to combine, such as text clustering and semantic similarity. We find that using clustering leads to the most competitive results on novel domains. 
\end{abstract}

\section{Introduction}

\renewcommand*{\thefootnote}{\fnsymbol{footnote}}
\footnotetext{$\star$\scalebox{0.91}{Correspondence to \href{mailto:achron@cis.lmu.de}{achron@cis.lmu.de} or \href{mailto:jessed@allenai.org}{jessed@allenai.org}}}
\footnotetext{$\dag$\scalebox{0.91}{Work done during an internship at Allen AI}}
\renewcommand*{\thefootnote}{\arabic{footnote}}

Large LMs are pre-trained using massive amounts of data in a self-supervised way \cite{peters-etal-2018-deep,devlin-etal-2019-bert,liu2019roberta, radford2019language} and obtain general-domain knowledge. In order to adapt them to a new domain, continuing training using in-domain data has been shown to be helpful \cite{han-eisenstein-2019-unsupervised, lee2020biobert, gururangan-etal-2020-dont}.
To avoid fine-tuning all parameters, efficient methods such as domain-specific mixtures-of-experts \cite{gururangan-etal-2022-demix} and hierarchical domain adapters \cite{chronopoulou-etal-2022-efficient} have been proposed.  Additional in-domain gains can be obtained using weight-space averaging \cite{pmlr-v162-wortsman22a, fisheravg}. Motivated by this, we propose using weight-space averaging at test time to improve performance on \textit{novel} domains \textit{without extra training}. 

\begin{figure}[!t]
	\centering
	\includegraphics[width=\columnwidth, page=1]{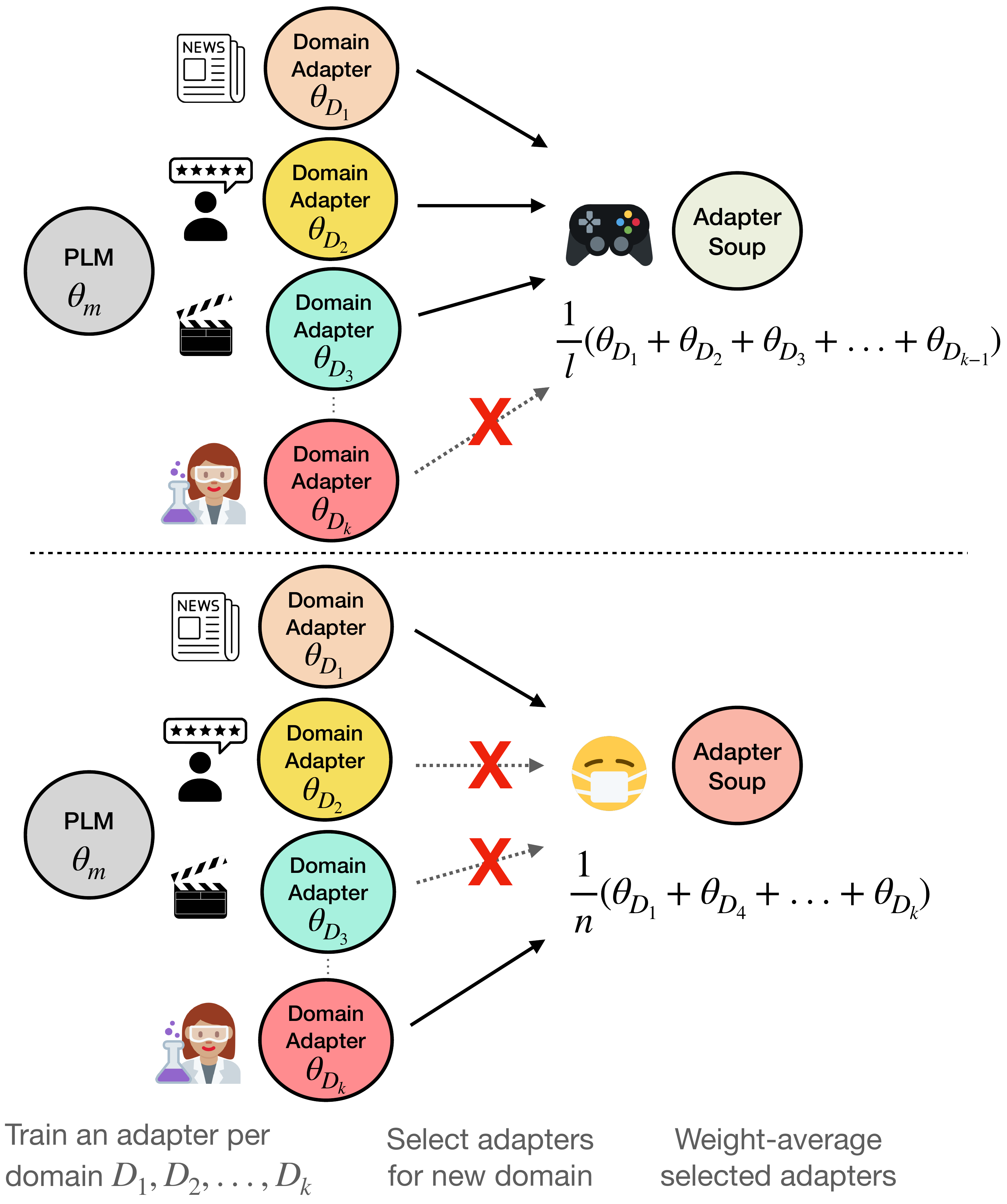}
        	\caption{Illustration of AdapterSoup. Starting from the same random seed, an adapter is trained for each domain (\textit{domain adapter}) on top of a PLM. AdapterSoup averages the weights of the adapters that are \textbf{most related} to the new domain to improve out-of-domain performance of a PLM at \textbf{test time}. The inference cost is independent of the number of adapters ($l$ or $n$) used. 
        }
	\label{fig:adaptersoup}
\end{figure}

 Our approach, AdapterSoup, ensembles adapters in the \textit{weight space}
  to improve performance on novel domains 
  at test time without parameter updates. 
 To this end, we train adapters on top of a PLM, each in a different domain. We compare several methods for selecting which adapters to use for each novel domain at test time and propose weight-space averaging models selected using text clustering. We find that AdapterSoup improves performance on novel domains. We also explore weight averaging adapters trained in the \textit{same} domain, each with a different hyper-parameter configuration, and find that combining models trained with a low learning rate provides competitive in-domain scores, while averaging models trained with high learning rates performs similarly to a general-purpose PLM on novel domains. 
 
 Our contributions are the following: \textbf{\textit{1})} We propose combining domain-adapted PLMs at inference time using adapters. Our approach leads to consistent gains in novel domains. We compare several methods for choosing the models of the AdapterSoup, concluding that text clustering provides the best performance across all domains. 
\textbf{\textit{2})} 
We perform weight-space averaging of PLMs adapted to the same domain with varied hyper-parameters using adapters. We find that we can obtain competitive in-domain scores but also preserve the generalization ability of a PLM.

\section{Proposed Approach}

\noindent\textbf{Problem Statement.} 
Assuming we have a PLM adapted to $k$ domains $D_1, ..., D_k$, we want a model that performs well in a novel domain $D_{k+1}$ without training more parameters. We use the provenance of a piece of text (that is, the \textit{website} from which the text was scraped) as a proxy for \textit{textual domain}. This follows \citet{chronopoulou-etal-2022-efficient,gururangan-etal-2022-demix}. 

If we assume that we have a PLM fine-tuned on a single domain $D_i$ with different hyper-parameters, we want to combine the fine-tuned models in order to both obtain good in-domain performance and preserve the generalization ability of the PLM to novel domains. 

\subsection{Cross-Domain AdapterSoup} 

An illustration of the cross-domain AdapterSoup is provided in Figure \ref{fig:adaptersoup}. Let $f(x, \theta_m)$ be a PLM with input data $x$ and parameters $\theta_m$ $\in$ $\mathbb{R}^d$. We add adapters with a parameter initialization $\theta_\alpha$. While in this work we parameterize $\theta_\alpha$ with adapters, our method is general and could be extended to other efficient fine-tuning methods.
We only fine-tune the adapters, without updating the parameters $\theta_m$ of the PLM, for language modeling using cross-entropy loss. 
Let us assume that $\theta$ = $\text{FineTune}(\theta_m,\theta_\alpha, \phi, D)$ denote the parameters obtained by fine-tuning a PLM with adapters in a domain $D$, using  hyper-parameters $\phi$.   

Let $\phi$ be a fixed hyper-parameter configuration. We vary only the \textit{textual domain}. 
We first train $k$ different adapters, one for each of the training domains. Then, we combine their weights: 
\setlength{\abovedisplayskip}{2pt}
\setlength{\belowdisplayskip}{2pt}
\begin{align}
AdapterSoup(x) = f(x, \frac{1}{l} \sum_{i=1}^{l} \theta_{i}),
\label{equation:1}
\end{align}

\noindent i.e., we use the average of the parameters of $l$ fine-tuned models, selected by one of the methods described in $\S$\ref{ssec:selection} ($l<=k$). If $l=k$, this model is a \textit{uniform soup} \cite{pmlr-v162-wortsman22a}.

\subsection{Single-Domain AdapterSoup} 
In this setup, we want to learn an LM that performs well in a single training domain $D$, while \textit{maintaining} the performance of the initial PLM $\theta_m$ in novel domains. To this end, we train adapters on the same domain, varying the \textit{hyper-parameter configuration}. Each of the $n$ models is optimized with different hyper-parameters  $\phi_i$, with $i \in 1,..., n$. 
We then compute the weight-space average following Equation \ref{equation:1}, with $l=3$.
This is similar to logit ensembling, but only adds to the PLM the inference cost of a single adapter, while the added inference cost of logit ensembling scales linearly with the number of adapters.

\subsection{Model Selection for AdapterSoup}
\label{ssec:selection}

In this section we describe two methods for selecting the combination of models to create our AdapterSoup (by weight-space averaging) which will be evaluated on a \textit{novel} domain $D_{k+1}$. Following standard practice \citep{gururangan-etal-2022-demix,btm-suchin} we use a small amount of  validation data from the novel domain $D_{k+1}$ for each of the below approaches. We note that we keep the test data unseen and only use it to perform our test-set evaluations.

\noindent \textbf{Sentence similarity.}
We use pretrained sentence-BERT \cite{reimers-gurevych-2019-sentence}, an approach that modifies BERT \cite{devlin-etal-2019-bert}
using siamese and triplet networks \cite{siamese} to obtain sentence embeddings.
We compute the embeddings for $100$ sentences from each of the training domains $D_1, ..., D_k$, plus the novel domain $D_{k+1}$. Then we compute the average cosine similarity between each of $D_1,..., D_k$ and $D_{k+1}$.
We add up to 5 adapters to the AdapterSoup in order of highest cosine similarity (only considering models trained on domains with cosine similarity greater than $0.15$ to $D_{k+1}$).  We experimented with several values to define the threshold ($3, 5, 10, 15$). We did not observe significant improvement when scaling up from $5$ to $10$  adapters and for that reason, we used up to $5$ adapters in each AdapterSoup.

\noindent \textbf{Domain clustering.} 
Our domain clustering approach follows \citet{aharoni-goldberg-2020-unsupervised}. We encode $100$ sequences from each of the training domains
using a PLM and fit a Gaussian Mixture Model (GMM) with $21$
components (equal to the number of training domains), which gives us a domain clustering.
We then use $100$ 
sequences from our held-out set (not used for test-set evaluation) and find which clusters they are closest to.
We add up to $5$ adapters to the AdapterSoup in order of which clusters the most held-out domain text is mapped to. If at least 10\% of the sequences of the  $D_{k+1}$ is mapped to the cluster of $D_i$, we add the model trained on $D_i$ to the AdapterSoup. 

\noindent \textbf{In-domain.}
To select the models that perform best in-domain, we exhaustively combine all models trained on a single textual domain (in this case, text found in the website \textit{booking.com}), using combinations of size 3. Each model has been trained with a different hyper-parameter configuration. Specifically, we vary the learning rate and data order. We compare them to the best-performing single model per domain and to a uniform soup.

\begin{table*}[]
\resizebox{\textwidth}{!}{
\begin{tabular}{lrrrrrrrrrr|r}
\toprule
 & \multicolumn{10}{c}{\textbf{10 Evaluation Domains}}               &         \\
  \textbf{Method}  & \textbf{reuters} & \textbf{techcrunch}&  \textbf{fastco} &  \textbf{nme} &  \textbf{fool} & \textbf{inquisitr} & \textbf{mashable} & \textbf{tripadv} & \textbf{ncbi} & \textbf{yelp} & \multicolumn{1}{c}{\textbf{Avg.}} \\
  \midrule 
GPT-2 (zero-shot)   &  21.5        &  27.7       & 27.9      &  28.2      & 23.8     &  22.4     & 27.1     & 40.4   & 20.7       & 36.2      & 27.6 \\ \hline

Single Adapter Chosen Using:   &             &         &       &        &      &       &      &     &        &     & \\
\hspace{3pt} - Sentence similarity & 18.9 & 22.0 & 22.0 & 23.1 & 22.9 & 18.4 & 25.3 & 37.0 & 18.2 & 49.4 & 24.4 \\
\hspace{3pt} - Clustering & 17.6 & 22.4 & 24.0 & 21.1 & 23.3 & 18.7 & 23.6 & 37.7 & 18.2 & 44.3 & 24.0 \\

AdapterSoup (Weight-space average):     &             &         &       &        &      &       &      &     &        &     & \\
 \hspace{3pt} - Uniform   & 18.2  & 23.1 & 22.9 &  22.2 & 22.4& 18.4 &  23.1 & 37.0 & 19.1 & 36.2 & 24.3 \\
 \hspace{3pt} - Sentence similarity & 17.6  & 22.0 & \textbf{21.3} &  \textbf{20.7}  & \textbf{22.2} & 18.4 & 22.4   & 36.2 &       \textbf{17.6}  & 35.2  & 23.4 \\
 \hspace{3pt} - \textbf{Clustering}  & \textbf{17.3} & \textbf{21.8} & \textbf{21.3} & 21.1 & \textbf{22.2} & \textbf{17.8} & \textbf{22.2} & \textbf{34.8} & \textbf{17.6} & \textbf{34.8} & \textbf{23.1} \\ 
\hline

Oracle     &             &         &       &        &      &       &      &     &        &     & \\
\hspace{3pt} - Best adapter per domain &   17.6 & 22.0 & 21.5 & 21.1  & 22.9 & 17.8 & 22.2 & 37.0 & 18.2  & 35.9  & 23.6 \\
 \hspace{3pt} - Clustering + 2 best   &  17.3 & 21.8 & 21.3 & 20.7 & 22.0 & 17.6 & 22.0 & 33.4 & 17.6    & 33.4 & 22.7 \\ 

 \midrule
Hierarchy adapter  & 16.4 & 20.1 & 20.1 & 20.1 & 22.2 & 16.4 & 22.2 & 33.1  & 18.2 & 34.5  & 22.3 \\  \bottomrule   
\end{tabular}
}

\caption{Perplexity ($\downarrow$) scores on $10$ evaluation domains.
All single adapter and AdapterSoup experiments have the same inference cost; bold indicates the best perplexity for each novel domain and best average. We find that AdapterSoup using clustering as a selection method on average leads to the best out-of-domain performance.} 
\label{table:crossdom}

\end{table*}

\section {Experimental Setup}
\noindent \textbf{Datasets.} We assume that text found in a specific website (e.g., \textit{tripadvisor}) can be used as a proxy of a textual domain. We use $21$ training domains and $10$ evaluation domains (text from $21$ and $10$ websites accordingly) from the released version \cite{dodge-etal-2021-documenting} of C4 \cite{JMLR:v21:20-074} (details in the Appendix).  We hypothesize that the variety of training domains plays an important role in this setting. We randomly sampled domains that belong to the $100$ high-resource domains of C4, but further work could consider using M$2$D$2$ \cite{m2d2}, a multi-domain language modeling dataset released concurrently to this work.

\noindent \textbf{Model Architecture.} We use GPT-2 \cite{radford2019language}; specifically, we use a publicly available pretrained checkpoint of the small version, i.e., \texttt{gpt2} from the HuggingFace library \cite{wolf-etal-2020-transformers}. We add an adapter to each Transformer \cite{NIPS2017_3f5ee243} layer after the feed-forward layer. We train only the adapters for language modeling in each training domain. The adapters follow the \citet{bapna-firat-2019-simple} architecture and have bottleneck size $64$. For the cross-domain AdapterSoup, we train all models with an initial learning rate $1e$-$4$. For the single-domain AdapterSoup, we use different learning rates and data seeds shown in the Appendix. 

\section{Results}
Results are presented in Table \ref{table:crossdom}. For each experiment, we  evaluate both perplexity and efficiency.

\subsection{Cross-domain} 
As a first baseline, we use \emph{GPT-2 (zero-shot)}, without further training or additional parameters. 
This has worse perplexity than all other approaches but is most efficient at inference.

\noindent\textbf{Single Adapters.}
We then evaluate \emph{Sentence similarity} and \emph{Clustering} in the scenario where only a single adapter is chosen using each approach (this can be thought of as a soup of size 1).
This is an evaluation of how well these two approaches measure similarity between the novel domain $D_{k+1}$ and the training domains; this baseline shows the performance of a single model which can be directly compared to AdapterSoups.
Both approaches are significantly better than GPT-2 (zero-shot), and \emph{Clustering} outperforms \emph{Sentence similarity}, suggesting it is better at identifying related domains.

\noindent\textbf{AdapterSoup.}
We evaluate three types of \emph{AdapterSoup} which differ only in how the models added to the soup are selected.
All three are equally as efficient at inference as using a single adapter. 
\emph{Uniform} is a uniform soup (weight-averaging all trained models).
This performs worse than all approaches except GPT-2 (zero-shot); 
we hypothesize that it performs worse due to negative interference between  adapters trained on unrelated domains.
Using \emph{Sentence similarity} as described in $\S$\ref{ssec:selection} leads to marginally better scores than the single-best adapter per domain, indicating even relatively naively-created soups can outperform the best (oracle) single model. On 8/10 novel domains, the sentence similarity AdapterSoup outperforms the single adapter chosen by Sentence similarity, indicating that the soup leads to better performance.
Next, using \emph{Clustering} as described in $\S$\ref{ssec:selection} leads to perplexity improvements in 8/10 novel domains compared to sentence similarity, indicating that the method for selecting models for the soup has a large impact.
On 9/10 novel domains, the Clustering AdapterSoup outperforms the single adapter chosen by clustering, indicating that our approach leads to better performance.

\noindent\textbf{Oracle Experiments and Larger Models.}
\emph{Best adapter per domain} shows the performance of the single-best adapter on each novel domain. This is the upper bound for a single adapter, and we see that our \emph{Single Adapter Chosen Using Clustering} matches these scores on $3/10$ novel domains, and is close on the rest, suggesting the clustering approach is reasonably good.
\emph{Clustering + 2 best} shows the performance of adding the two (oracle) best models to our AdapterSoup made by clustering; our clustering approach is close to these scores, but there is room for future work on better choosing models for the AdapterSoup. 
\emph{Hierarchy adapter} is taken from \citet{chronopoulou-etal-2022-efficient}, and is less efficient in terms of both data and parameters.

\begin{table}
\centering
\small
\resizebox{0.9\columnwidth}{!}{%

\begin{tabular}{lrr}
\toprule
\textbf{Novel Domain} $i$  & \textbf{Sentence Sim.} & \textbf{Clustering}  \\ \midrule 

\textbf{tripadvisor} & booking & booking \\ 
  & insiderpages & insiderpages \\ 
&  & lonelyplanet \\ \midrule
\textbf{ncbi} & journals & journals \\ 
  & frontiersin & frontiersin \\ 
  & springer & springer \\  \midrule 
\textbf{reuters} & csmonitor & dailymail \\ 
 & wired & express \\ 
& entrepreneur &  \\

\bottomrule 
\end{tabular}}
\caption{Domains of models selected for the  AdapterSoup using either sentence similarity or  clustering. The clustering method seems to more accurately match each novel domain to training domains that are similar to it.}

\label{table:wsadomains}
\end{table} 

\noindent\textbf{Selecting Models for the Soup.}
We qualitatively compare the selection methods for choosing adapters to include in the AdapterSoup for 3 novel domains in Table \ref{table:wsadomains}. In the case of \textit{tripadvisor}, 2/3 domains \textit{Sentence similarity} and \textit{Clustering} select are identical, while for \textit{ncbi} (science domain) both methods select the same domains. When selecting domains similar to \textit{reuters} (news), clustering seems to find a good match, choosing news domains. However, \textit{Sentence similarity} selects domains that are not quite as related to the novel domain. \textit{Reuters} contains heterogeneous data, so the average cosine similarity on the sentence level is not a suitable metric to find related domains. 

\begin{table}
\centering
\small
\resizebox{\columnwidth}{!}{%

\begin{tabular}{lr|rrr}
\toprule

    & \textbf{booking}  & \textbf{frontiers} & \textbf{journals}   & \textbf{yelp}  \\
 & ID & OOD & OOD  & OOD \\ \midrule 

  GPT-2 (zero-shot) & 29.7 & 22.2 & 24.5  & 36.2 \\  \midrule 
  Best single adapter& 10.2 & 27.7 & 30.3  & 49.4 \\
AdapterSoup: &  & &&  \\ 
\hspace{4pt}- lr $7e$-$3$ & 27.7 & 23.3 & 24.8  & 37.7 \\
\hspace{4pt}- lr $4e$-$3$ & 24.5 & 23.8 & 25.5 & 39.6 \\
\hspace{4pt}- lr $1e$-$3$ & 11.5 & 24.0 & 26.3  & 42.5 \\
\hspace{4pt}- lr $5e$-$4$ & 10.0 & 26.3 & 29.1  & 47.5 \\
\hspace{4pt}- lr $1e$-$4$  & 10.4  & 27.4 & 30.0 & 48.9 \\

Best AdapterSoup:  &  & &&  \\  
   \hspace{4pt}- in-domain  & \textbf{10.0}  & 26.3 & 29.1& 47.5 \\ 
   \hspace{4pt}- out-of-domain & 26.8  & \textbf{22.9} & \textbf{24.5} & \textbf{37.3} \\ \midrule 
  Logit ensemble & 9.2&    25.0 & 27.7&  47.7\\  \bottomrule

\end{tabular}}
\caption{Perplexity scores in- and out-of-domain (respectively ID and OOD) of models trained on \textit{booking.com}. Low learning rates lead to good in-domain scores, while high learning rates improve the out-of-domain performance.}

\label{table:singledomain}
\end{table}

\subsection {Single-domain} 
In this section we evaluate how models trained on the same domain can be combined into an AdapterSoup.
We train a set of models using adapters on \textit{booking.com} by varying the data order and the learning rate (see Appendix~\ref{app:single_adapter}, note our experiments kept the initialization of each adapter fixed), then evaluate all combinations of adapters of size $3$, and evaluate the performance of the AdapterSoup both in-domain (\textit{booking.com}) and on $3$ held-out domains.
We explore this controlled setting to better understand the setup described in \citet{pmlr-v162-wortsman22a}, who also noted that the learning rate is important; their experiments indicated that smaller learning rates led to better model soups.

Our experiments in Table~\ref{table:singledomain} show a more nuanced result: AdapterSoups made from adapters trained with small learning rates ($5$e-$4$) performed best in-domain (confirming the result from \citealp{robustzeroshot}), but AdapterSoups made from adapters trained with larger learning rates ($7$e-$3$, $4$e-$3$, and $7$e-$4$) generalize better to novel domains.
The number of updates for each adapter is the same, and they all have the 
 same initialization, so we hypothesize that AdapterSoups made from small learning rates act similarly to averaging across steps in gradient descent, leading to a model that is closer to a local optimum.
As for why larger learning rates leads to better generalization to novel domains, we hypothesize that each model in the AdapterSoup travels a farther distance from the initialization, leading to learning somewhat more diverse representations. We leave further exploration to future work.

\section{Related Work} 
As training large models from scratch has a severe computational and environmental cost \cite{strubell-etal-2019-energy, co2}, efficient methods such as mixtures-of-experts (MoE) \cite{moe-shazeer,switchtransformer,artetxe2021efficient}, adapters \cite{rebuffi,houlsby, pfeiffer-etal-2020-mad}, 
and
LoRA layers \cite{lora} have recently been proposed. Both adapters and MoEs have shown to work well for domain adaptation \cite{cooper-stickland-etal-2021-multilingual,gururangan-etal-2022-demix,chronopoulou-etal-2022-efficient}. The hierarchy adapter \cite{chronopoulou-etal-2022-efficient} outperforms our approach but is significantly more expensive. It adds a training cost of $4Ld_{\textrm{model}}dT$ (following \citealp{Kaplan2020ScalingLF}) over the cost of running GPT-2 for a model with $L$ layers,  dimension $d_\textrm{model}$, adapter bottleneck size $d$, average tree depth $T$ ($T=8$ in the hierarchy adapter paper), while AdapterSoup needs $4Ld_{\textrm{model}}d$ flops. As a result, training the hierarchy adapter is a factor of $T$ slower than our approach. At inference time, the hierarchy adapter activates 2 paths in the tree and invokes a cost $4Ld_{\textrm{model}}dT \times 2$, i.e., inference is a factor of $2T$ slower than our approach.

Averaging \textit{weights} of models independently fine-tuned on the same task \cite{pmlr-v162-wortsman22a} has shown to improve in-domain  performance. \citet{fisheravg} weight-average fine-tuned PLM models using Fisher merging to avoid intermediate task training and then perform downstream fine-tuning. \citet{adamix} fine-tune MoEs using adapters on a downstream task and average their weights at test time. Our paper, however, focuses on improving test-time scores of a model on \textit{novel} domains. 

\citet{wang-etal-2021-efficient-test} improve performance in an unseen (target) language by ensembling the source language adapter and  language adapters similar to the target language. This approach uses weighted ensembling of the \textit{outputs} of adapters, whereas we ensemble the \textit{weights} of the adapters. AdapterSoup has the inference cost of a single adapter, while \citet{wang-etal-2021-efficient-test} require inference time that scales linearly to the number of adapters.

Contemporaneous work  \cite{btm-suchin} also explores performance in novel domains using weight averaging, but averages expert language models (entire models), while we average adapters. 

\section{Conclusion}

A PLM can be adapted to new domains using adapters. However, this requires training a new set of adapters for each domain. We propose a method based on weight-space averaging of adapters selected using text clustering. Our approach improves performance on novel domains without updating parameters or increasing the inference cost. Future work could explore more sophisticated selection methods to try to match the performance of the oracle experiments.

\section*{Limitations}
The conclusions we draw in this work about how our approach compares to other approaches (e.g., our baselines) are only supported by evidence on the task of language modeling, with textual domains taken from the C4 dataset. We expect such results to hold more generally, but do not have experimental evidence to support any other scenarios.
As with all work on language modeling, the models we have trained could be used to generate language, but we do not have evaluations of generated text (e.g., on fluency, factuality, or other common metrics used to evaluate generated language).
Our paper focuses on using adapters; while we expect similar approaches to work for other types of models, we only have evidence to support AdapterSoup working for adapters.

\section*{Acknowledgements}
We thank Ayyoob Imani for feedback on the final version of the paper and Jonas Pfeiffer for helpful discussions. We also thank Mitchell Wortsman and Ludwig Schmidt for preliminary comments on the first version of this idea.

\bibliography{custom,anthology}
\bibliographystyle{acl_natbib}
\clearpage

\newpage
\appendix

\section{Appendix}
\label{sec:appendix}
    \subsection{Training details} We build our code using PyTorch \cite{pytorch} and the HuggingFace library \cite{wolf-etal-2020-transformers}. Each model is trained on a single NVIDIA A100 GPU with 40GB of RAM, batch size 64 and gradient accumulation over 5  steps.  We train each model for 20 epochs, without using early stopping.  We compute semantic similarity using  sentence-transformers\footnote{\url{https://github.com/UKPLab/sentence-transformers}} and a publicly available pretrained model.\footnote{\url{huggingface.co/sentence-transformers/all-mpnet-base-v2}}

We noticed from preliminary experiments that the choice of random seed is important when averaging weights of domain adapters. We empirically found that averaging domain adapters initialized from different random seeds led to poor performance of AdapterSoup. We suggest initializing the adapters from the same random seed in order to effectively combine adapters trained on various domains. 
 
\subsection{Dataset sizes}  We use textual corpora from 30 of the 100 most high-resource internet domains of C4. \textit{yelp.com} is the only internet domain we used which comes from a different source.\footnote{\url{https://www.yelp.com/dataset}} The sizes of the training domains are shown in Table \ref{table:trainingcorpora}, while the sizes of the evaluation domains are shown in Table \ref{table:evalcorpora}.
\begin{table}[h]
\centering
\small
\resizebox{\columnwidth}{!}{%

\begin{tabular}{r|lr}
\toprule
\textbf{Ind} & \textbf{Training Domain}   & \textbf{Train (Eval.) Tokens} \\ \midrule 

 1& dailymail.co.uk    &25M (3M) \\ 

2& wired.com            & 18M (2M) \\
3& express.co.uk        & 16M (2M)   \\ 

 4& npr.org              & 25M (3M)   \\ 
5 & librarything.com     & 3M (500K)     \\ 
6 & instructables.com    & 25M (3M)       \\ 
7 & entrepreneur.com     & 16M (2M)   \\ 
8 & link.springer.com        & 28M (4M) \\
9 & insiderpages.com     & 8M (1M)       \\ 
10 &ign.com              & 10M (1M)          \\
11 & eventbrite.com       & 11M (1M)    \\
12 & forums.macrumors.com     & 22M (3M)   \\ 
 13 & androidheadlines.com & 14M (2M)         \\
14 &  glassdoor.com        & 4M (500K)  \\
 15 &   pcworld.com          & 14M (2M)             \\
 16 &  csmonitor.com        & 23M (3M)             \\
 17 & lonelyplanet.com         & 6M (1M)     \\
 18 &  booking.com        &  30M (4M)        \\
 19 &  journals.plos.org        & 53M (6M)         \\
 20 &  frontiersin.org        &  38M (6M)         \\
 21 &  medium        & 22M (3M)         \\
\bottomrule 
\end{tabular}}
\caption{Sizes of training corpora. We fine-tune GPT-2 using adapters on each of these domains. We perform weight-averaging of these 21 domain-adapted LMs.}

\label{table:trainingcorpora}
\end{table}

\begin{table}[t]
\centering
\small
\resizebox{\columnwidth}{!}{%

\begin{tabular}{r|lr}
\toprule
\textbf{Ind} & \textbf{Novel Domain}   & \textbf{Train (Eval.) Tokens} \\ \midrule 

1 & reuters.com   & 17M (2M)\\ 
2 & techcrunch.com   & 13M (2M)\\ 
3 & fastcompany.com   & 14M (2M) \\ 
4 & nme.com    & 5M (1M) \\ 
5 & fool.com    & 34M (4M) \\ 
6 & inquisitr.com   & 13M (2M) \\ 
7 & mashable.com    & 14M (2M) \\ 
8 & tripadvisor.com   & 7M (1M) \\ 
9 & ncbi.nlm.nih.gov    & 23M (3M) \\ 
10 &  yelp.com  & 684M (20M)   \\

\bottomrule 
\end{tabular}}
\caption{Sizes of held-out corpora. }

\label{table:evalcorpora}
\end{table} 

\subsection{Single-domain AdapterSoup}
\label{app:single_adapter}
We present the hyper-parameters we tried in Table \ref{table:hyperparams}. In this setup, we computed in- and out-of-domain scores for 455 different combinations (there are 15 models and computed all AdapterSoups of size 3). The trend we observed is that higher learning rates improved results out-of-domain, while lower learning rates provided the best in-domain scores.

\begin{table}[t]
\centering
\small
\resizebox{0.8\columnwidth}{!}{%

\begin{tabular}{lr}
\toprule
Hyper-parameter & Value \\ \midrule 
\multirow{2}{*}{learning rates}  & $7e$-$3$, $4e$-$3$ \\ 
&  $1e$-$3$, $5e$-$4$, $1e$-$4$ \\ 

random seed & 1, 2, 3 \\ \bottomrule

\end{tabular}}
\caption{Hyper-parameters for single-domain AdapterSoups. We exhaustively compute the AdapterSoup for every combination of 3 models in this set.}

\label{table:hyperparams}
\end{table} 

\subsection{Cross-domain AdapterSoup}
We present in Table \ref{table:allscoresmodel1} the evaluation scores of each of the single adapter models. Each adapter has been trained in a different training domain (column 1), and evaluated in 10 novel domains. 

\begin{table*}[]
\resizebox{\textwidth}{!}{
\begin{tabular}{lrrrrrrrrrrr} \toprule
                 & \multicolumn{10}{c}{\textbf{Evaluation Domains}}                                    & \multicolumn{1}{l}{}                                    \\
\textbf{Training Domain}   & \multicolumn{1}{l}{{\textbf{reuters}}} & \multicolumn{1}{l}{{\textbf{techcrunch}}} & \multicolumn{1}{l}{\textbf{fastco}} & \multicolumn{1}{l}{{\textbf{nme}}} & \multicolumn{1}{l}{{\textbf{fool}}} & \multicolumn{1}{l}{{ \textbf{inquisitr}}} & \multicolumn{1}{l}{{\textbf{mashable}}} & \multicolumn{1}{l}{{\textbf{tripadv.}}} & \multicolumn{1}{l}{\textbf{ncbi}} & \multicolumn{1}{l}{{\textbf{yelp}}} & \multicolumn{1}{c}{{\textbf{Avg}}} \\ \midrule
dailymail       & 17.6 & 23.6 & 24.0 & {21.1} & 23.3 & {18.4} & 23.6 & 39.6 & 20.5 & 44.3 & 25.6    \\
wired       & {18.0} & {22.0} & 21.5 & 22.0 & 22.9 & 18.2 & 22.2 & 40.0 & 19.9 & 41.3 & 24.8     \\
express     & 19.5 & 25.8 & 26.0 & {22.6} & 25.8 & 20.1 & 26.3 & 42.9 & 23.3 & 48.9 & 28.1      \\
npr          & 20.1 & 25.5 & 25.0 & 27.7 & 23.3 & {20.5} & {23.6} &  42.1 & 21.1 & 42.9 &27.2    \\
librarything   & 19.5 & 24.5 & 24.0 & 24.8 & 23.6 & 19.7 & 24.8 & 38.9 & 21.1 & 39.3 & 26.0     \\
instructables & 20.5 & 25.5 & 25.5 & 25.5 & 24.5 & 20.5 & 25.5 & 40.0 & 21.1 & 41.7 & 27.0    \\
entrepreneur   & {18.2} & {22.4} & {22.0} & 22.6 & {22.9} & 18.4 & 23.1 & 40.9 & 21.1 & 43.4 & 25.5    \\
springer       & 19.7 & 25.0 & 24.5 & 24.5 & 25.3 & 19.9 & 26.8 & 42.9 & {18.4} & 43.8 & 27.1    \\
insiderpages  & 23.1 & 28.8 & 29.1 & 32.1 & 25.5 & 23.1 & 27.9 & {37.7} & 23.3 & 35.9 & 28.7       \\
ign         & 18.9 & 23.8 & 23.6 & 22.6 & 23.3 & 18.7 & 23.6 & 40.9 & 21.1 & 39.6 & 25.6       \\
eventbrite   & 19.1 & 24.3 & 23.8 & {23.1} & 24.3 & {19.3} & 25.0 & 39.6 & 20.9 & {41.7} & 26.1       \\
macrumors    & 20.3 & 26.0 & 26.3 & 26.3 & 24.5 & 20.9 & 25.5 & 41.3 & 22.4 & 43.4 & 27.7        \\
androidheadlines & 20.7 & {24.8} & 25.8 & 26.0 & {24.5} & 20.1 & {25.3} & 44.7 & 22.6 & 42.9 & 27.8   \\
glassdoor      & 20.7 & 26.0  & {25.8} & 27.7 & {24.8} & 21.1 & 26.8 & 42.5 & 22.0 & 42.5 & 28.0         \\
pcworld    & 18.7 & {22.6} & 22.9 & 23.6 & {23.1} & 18.7 & {23.1} & 42.1 & 21.5 & 42.9 & 25.0     \\
csmonitor      & {18.9} & 24.0 & 23.8 & 24.0 & 23.6 & {18.9} & 23.8 & 41.3 & 21.5 & 43.4 & 26.3         \\
lonelyplanet     & 20.7 & 26.0 & 25.8 & 25.0 & 25.3 & 20.7 & 26.6 & 40.4 & 22.6 & {42.9}  & 27.6     \\ 
booking     & 27.4 & 33.4 & 33.1 & 35.9 & 31.5 & 27.4 & 35.5 & {37.0} & 30.6 &{49.4} & 34.1        \\
journals     & 21.3 & 26.8 & 26.0 & 27.4 & 26.0 & 21.5 & 28.2 & 46.1 & {18.2} & 46.5 & 28.8        \\   
frontiersin     & 21.1 & 26.8 & 25.5 & 27.7 & 26.0 & 27.7 & 26.0 & 45.6 & {19.3} & 46.5 & 29.2   \\
medium     & 17.8 & 22.2 & {21.8} & 21.3 & 25.0 & 17.8 & 25.3 & 39.3 & 19.9 & 43.4 & 25.4       \\  
    \bottomrule 
 \end{tabular}}
\caption{We show the performance of each trained adapter (for the cross-domain setting) on the 10 evaluation domains. Each model has been trained for language modeling with an initial learning rate $1e-4$ for 20 epochs. }
\label{table:allscoresmodel1}

\end{table*}

\end{document}